\pdfoutput=1
\documentclass[11pt]{article} 

\usepackage{acl}
\usepackage{latexsym}
\usepackage{listings}
\usepackage[T1]{fontenc}
\usepackage{microtype,cuted,longtable}
\usepackage{times,multicol}
\usepackage{url,tcolorbox}
\usepackage{paralist,multicol,xtab}
\usepackage{graphicx}
\usepackage{wrapfig}
\usepackage{fancyvrb}
\usepackage[inline]{enumitem}
\usepackage{amssymb,multirow}
\usepackage{natbib}
\usepackage{booktabs,fvextra }
\usepackage{comment}
\usepackage{subcaption}
\usepackage{amssymb}
\usepackage{verbatim}
\usepackage{longtable}
\usepackage{tikz}
\usepackage{glossaries}
\usepackage{array}
\usepackage{enumitem,verbatimbox}
\usepackage{inconsolata}
\usepackage[a4paper,margin=2.5cm]{geometry}
\usepackage{xspace}

\newcommand{\fig}{Fig.\xspace}
\newcommand{\tab}{Tab.\xspace}
\newcommand{\critic}{$C=\{c_1,\dots, c_n\}$\xspace}
\newcommand{\criterion}{$c_i: \{\omega_j\}_{j=1}^m$}
\newcommand{\utility}{$U_t(s)=\{Q_i(s|c_i)\}_{i=1}^n$\xspace}
\newcommand{\qa}{\emph{QuantifierAgent}\xspace}
\newcommand{\ca}{\emph{CriticAgent}\xspace}
\lstdefinelanguage{json-like}{
  basicstyle=\ttfamily,
  numbers=left,
  numberstyle=\tiny\color{gray},
  numbersep=5pt,
  showstringspaces=false,
  breaklines=true,
  frame=single,
  morestring=[b]",
  stringstyle=\color{blue},
  morecomment=[l]{//},
  commentstyle=\color{green},
  morecomment=[s]{/*}{*/},
  morekeywords={content, role}
}

\newcounter{todocnt}

\title{Towards better Human-Agent Alignment: \\Assessing Task Utility in LLM-Powered Applications}

\author{
\textbf{Negar Arabzadeh}$^{1}$\thanks{\phantom{\_\_}Work done during an internship at Microsoft Research} \quad 
\textbf{Julia Kiseleva}$^2$ \quad 
\textbf{Qingyun Wu}$^3$ \quad 
\textbf{Chi Wang}$^2$ \quad \\
\textbf{Ahmed Awadallah}$^2$ \quad  
\textbf{Victor Dibia$^2$} \quad 
\textbf{Adam Fourney$^2$} \quad
\textbf{Charles Clarke$^1$} \\
$^1$Univerity of Waterloo\\
$^2$Microsoft Research\\
$^3$Pennsylvania State University\\
}

\begin{document}

\maketitle
\begin{abstract}

The rapid development in the field of Large Language Models (LLMs) has led to a surge in applications that facilitate collaboration among multiple agents to assist humans in their daily tasks. However, a significant gap remains in assessing whether LLM-powered applications genuinely enhance user experience and task execution efficiency. This highlights the pressing need for methods to verify utility of LLM-powered applications, particularly by ensuring alignment between the application's functionality and end-user needs.
We introduce AgentEval\footnote{\url{https://github.com/microsoft/autogen/blob/main/notebook/agenteval_cq_math.ipynb} provides an implementation for the math problems}, a novel framework designed to simplify the utility verification process by automatically proposing a set of criteria tailored to the unique purpose of any given application. This allows for a comprehensive assessment, quantifying the utility of an application against the suggested criteria. We present a comprehensive analysis of the robustness AgentEval for two open source datasets.

\end{abstract}

\section{Introduction}
\label{sec:intro}

The rapid development of open-source libraries~\cite{wu2023autogen,li2023camel} that aims to simplify the development of LLM-powered agentic solutions for various user-enteric tasks has led to the rapid growth of such applications~\cite{liang2023encouraging,hong2023metagpt,talebirad2023multi}. One of the long-lasting goals~\cite{winograd1972understanding} is the ability to seamlessly interact with humans in natural language to help end-users and to make their lives easier by assisting with their tasks from math tutoring to completing household tasks and so on. End users have expectations and requirements for a developed application that need to be met. This understanding is essential to assess the \emph{utility} it brings and, consequently, to further improve and align the application towards end-users' goals.

Directly evaluating agentic systems poses challenges as current approaches predominantly rely on end-to-end success metrics – essentially, whether the agent accomplishes tasks~\cite{shridhar2020alfworld, DBLP:journals/corr/abs-1912-01734,pmlr-v229-myers23a}. However, comprehending user interaction with an application involves much more than success alone~\cite{pmlr-v176-kiseleva22a, pmlr-v220-kiseleva22a, zhang2023creative}. Take math problems, for instance; it is not merely about the agent solving the problem. Equally significant is its ability to present solutions based on various criteria, including completeness, conciseness, and clarity of the explanation provided. In other words, in a code completion scenario, even an incomplete code suggestion can be useful when it provides substantial boilerplate code or proposes a framework to solve a task~\cite{dibia2023aligning}. Furthermore, success is not always clearly defined for every task. Knowing such criteria for an LLM-powered application and being able to quantify them is essential to verify whether user requirements are being satisfied, in other words, if the application brings utility to the end-users. Given the objective of verifying arbitrary applications, reliance on a benchmarking approach is untenable due to the expansive range of tasks requiring automation. A prerequisite is a scalable and flexible methodology capable of accommodating a diverse set of applications.

\begin{figure*}[!tb]
\centering
  \includegraphics[scale=0.5]{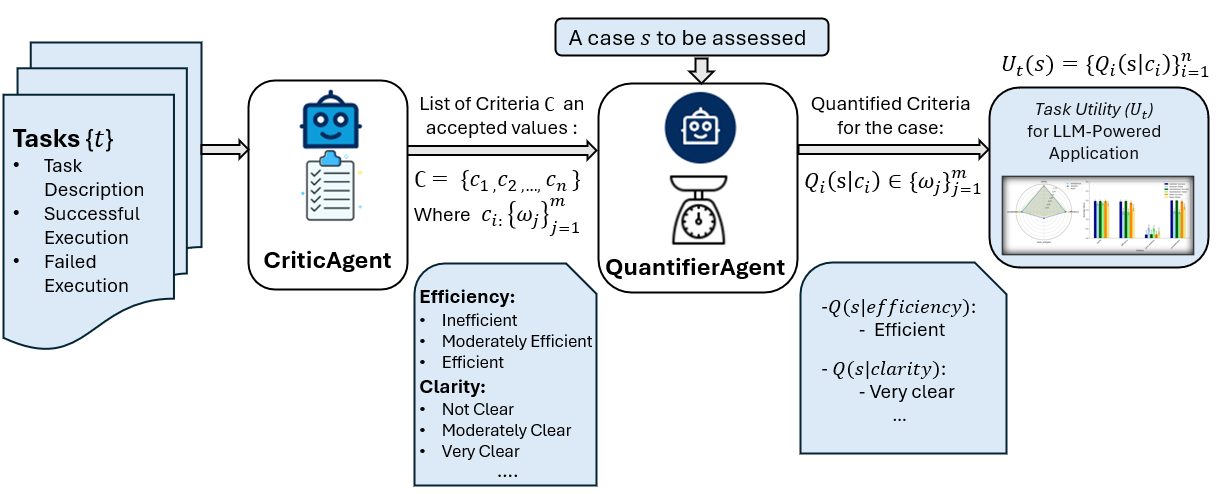}
\caption{An overview of the \emph{AgentEval} framework consists of two main components: (C) \ca, which learns a list of $n$ criteria (\critic) and suggested values for each criterion (\criterion), where $m$ is the number of suggested values, applicable to an arbitrary application that can be assessed by a domain expert; and (Q) \qa, which verifies a set of suggested criteria for a considered application and suggests a task utility for an end-user (\utility)}
\label{fig:agent-eval}
\end{figure*}

In this work, we aim to introduce the  AgentEval framework, a tool crafted to swiftly gauge the utility of LLM-powered agentic applications designed to help end-users accomplish their desired tasks. The goal of AgentEval is to assess the current alignment between application behavior and user goals, providing application developers with insights into how and what aspects of the current flow can be improved. AgentEval takes into account recent discoveries that have shown LLMs to emerge as a scalable and cost-effective alternative to human evaluations for open-ended tasks~\cite{li2023collaborative}. The \emph{AgentEval} is illustrated in \fig~\ref{fig:agent-eval}, which consists of two main agents executed consecutively. These agents are customizable, conversable, and can operate in various modes that employ combinations of LLMs, human inputs, and tools~\cite{wu2023autogen}\footnote{\url{https://github.com/microsoft/autogen}}:

\begin{itemize}[leftmargin=*, nosep]
    \item \ca suggests the list of criteria based on the task description and the suggested solutions, e.g. for math problems can be \emph{Efficiency} of the proposed solution and \emph{Clarity} of the proposed solution;
    \item \qa verifies how well the solution $s$ produced by an agentic system designed for task $t$ are performing for each criterion and returns the utility function, e.g. what is the Clarity level of the solution, not clear, moderately clear or very clear. 
\end{itemize}

We believe that the usage of \emph{AgentEval} can extend beyond the immediate verification of the current performance of LLM-powered applications. The framework can be employed over time to uncover new capabilities of the system and potential changes in task utility for the end user. The discovered utility function can be utilized to optimize the system towards user needs or system developer requirements, and this optimization can occur over time.

%%% List the conclusions
In summary, our main contributions are:
\begin{enumerate}[leftmargin=*, nosep, label=\textbf{C\arabic*}]
\item A definition of task utility that enables access to the possible requirements an end-user may have regarding an LLM-powered application and how well the application satisfies this list of criteria;
\item An introduction of \emph{AgentEval}, a novel framework that leverages LLM-powered agents as a scalable and cost-effective alternative to human evaluations to produce task utility through the collaboration of two agents: \ca proposes a list of criteria based on the task description and the successful and failed execution of the agent, and \qa assesses how well the list of criteria is supported by the current implementation of an application;
\item An in-depth analysis of the \emph{AgentEval} robustness on various tasks and datasets across different solutions that can be replicated for a newly unseen domain.
\end{enumerate}

\noindent
The remainder of this paper is organized as follows. Section~\ref{sec:rel-work} describes earlier work and background. We provide the motivation behind AgentEval and define the utility of the task in Section~\ref{sec:agenteval}. Section~\ref{sec:datasets} gives an overview of the datasets, namely MATH~\cite{hendrycks2021measuring} and ALFWorld~\cite{shridhar2020alfworld}, and the solutions for building applications powered by LLM utilized in our work. Section~\ref{sec:workflow} demonstrates our findings about applying AgentEval to assess the utility of the task for the selected datasets. Section~\ref{sec:results} presents an in-depth analysis of the robustness of AgentEval, namely \ca robustness (Section~\ref{sec:c-robust}), \qa robustness (Section~\ref{sec:q-robust}) and an automatic verification of \qa (Section~\ref{sec:q-verify}).

\section{Related Work}
\label{sec:rel-work}

We build on the streams of prior work.
First, we will discuss the list of benchmarks and approaches to evaluate general LLMs in general (Section~\ref{sec:llm-eval}). Second, we will present approaches to understand and predict user utility functions (Section~\ref{sec:sat-pred}).
Third, we will go over the current tendencies in using LLMs as evaluators in Section~\ref{sec:llm-evaluators}.

\subsection{LLM evaluation}

\label{sec:llm-eval}
There exists a significant volume of literature dedicated to assessing Language Model Models (LLMs), as evidenced by extensive research efforts~\cite{guo2023evaluating, ziyu-etal-2023-lens, chang2023survey, liang2023holistic}. LLMs have been evaluated from various aspects including, but not limited to, specialized LLMs~\cite{jin2019pubmedqa}, ethics and morality~\cite{hendrycks2020aligning}, safety and robustness~\cite{wang2023robustness}, and knowledge and reasoning~\cite{bian2023chatgpt}.
Additionally, recent developments include the introduction of intricate multi-modal benchmark datasets~\cite{mialon2023gaia, bang2023multitask}. Furthermore, there are attempts to evaluate LLMs as agents~\cite{liu2023agentbench}.

However, there is a lack of literature that focuses on the holistic verification of the utility of LLMs for end-users in solving their tasks, which we tackle in this work.

\subsection{User satisfaction prediction}
\label{sec:sat-pred}
Recent studies suggest that users interacting with various systems operate with specific utility functions in mind~\cite{li2020optimizing, azzopardi2018measuring, ahmadvand2022supporting}. Traditionally, metrics defining user satisfaction was built based on various at scale collected behavioral signals~\cite{kiseleva2014modelling} and they were tailored to specific applications, such as intelligent assistants~\cite{kiseleva2016predicting, kiseleva2016understanding}, web search engines~\cite{ williams2016detecting,williams2016your,williams2017does}, dialogue systems~\cite{see2019makes}, multi-turn conversations~\cite{li2021deus} and general-purpose personal assistants~\cite{kiseleva2017evaluating}.

\subsection{Using LLMs as evaluators}
\label{sec:llm-evaluators}
Moreover, there is a growing trend in utilizing LLMs as evaluates~\cite{chiang2023can,fu2023gptscore} for qualitative research~\cite{bano2023exploring} and adopting LLMs as proxies for human behavior~\cite{tjuatja2023llms, liu2023voices}. 
\citet{jain2023multi} studied efficacy of in-context learning based evaluators in evaluating zero-shot summaries written by LLMs. 
Notably, CoEval~\cite{li2023collaborative} has recently demonstrated the synergy between human evaluation and LLMs  in establishing evaluation criteria and conducting multi-dimensional evaluations for openended NLG tasks.

\medskip
\noindent
Building on top of these works, we propose a framework capable of assessing the utility of various LLM-powered applications at scale. This framework aims to align agentic systems with human preferences.

\section{Defining Task Utility}
\label{sec:agenteval}

\begin{figure}[!tb]
\centerline{%
  \includegraphics[clip, trim=6.1cm 3.5cm 11.2cm 4.4cm,scale=0.5]{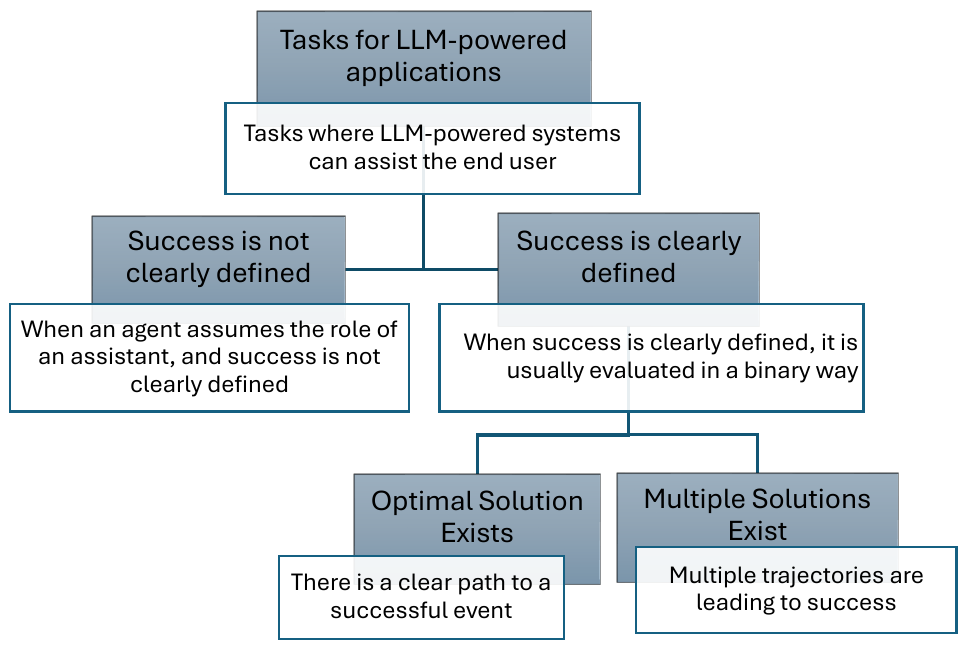}}
\caption{The taxonomy of task assessments based on optimal solutions existence}
\label{fig:task-taxonomy}
\end{figure}

It is important to begin by considering the categories of tasks around which we focus LLM-powered applications.
\fig~\ref{fig:agent-eval} outlines a taxonomy of target tasks for agentic systems in terms of success metrics. At the top level, tasks can be split into two main categories, where:
\begin{itemize}[leftmargin=*, nosep]
\item \emph{Success is not clearly defined} --- For these tasks, users utilize a system in an assistive manner, seeking suggestions rather than expecting the system to solve the task end-to-end. For example, a user might request the system to generate an email based on some user input. In many cases, this generated content serves as a template that the user will later edit. However, defining success precisely for such tasks is less well-defined. In case of online evaluation, while being expensive, we can ask users to what extend the the assistance was helpful. Although quantifying to what extent the help was useful is still challenging on its own, the problem becomes increasingly challenging when it comes to offline evaluation, or evaluation in novel scenarios before we have users.

\item \emph{Success is clearly defined} --- For these tasks, we can clearly determine whether a system solved the task or not. Consider agents that assist in accomplishing household tasks, where the definition of success is clear and measurable.
\end{itemize}
This second category can be further divided into two subcategories:
\begin{itemize} [leftmargin=*, nosep]
\item \emph{Success is clearly defined and an optimal solution exits} --- For these tasks, only one solution is possible. For example, if you ask your assistant to turn on the light, the success of this task is clearly defined, and there is only one way to accomplish it.
\item \emph{Success is clearly defined and multiple solutions exist} --- increasingly, we observe situations where multiple trajectories of agent behavior can lead to either success or failure. In such cases, it is crucial to differentiate between the various successful and unsuccessful outcomes. For example, when you ask the agent to suggest a food recipe or tell you a joke, you may define success as the food tasting good or the joke being funny, but perhaps the recipe should not be too expensive to prepare, and the joke should not be offensive.
\end{itemize}

\noindent
In our AgentEval framework, we are currently focused on tasks where success is clearly defined and multiple successful solutions may exist. 

Our previous research on assistive agents suggested that the most optimal way to obtain human judgments is to present humans with two agents side by side and ask for preferences~\cite{pmlr-v220-kiseleva22a}. In this setup of pairwise comparison, humans can develop a list criteria to explain why they prefer the behavior of one agent over another. For instance, `the first agent was faster in execution' or `the second agent moves more naturally'. Therefore, the comparative nature guided humans to come up with a list of criteria that helps to infer the utility of the task. With this idea in mind, we designed AgentEval (shown in Fig.~\ref{fig:agent-eval}), where we employ LLMs to help us understand, verify, and assess task utility for the multi-agent system. The AgentEval Framework employs two types of agents, namely:

\begin{itemize} [leftmargin=*, nosep]
\item The goal of a \ca is to suggest a list of criteria that can be used to assess task utility for end users. The critic is given a task description as well as a list of few successful and failed examples of the task execution; then it is able to return a list of criteria: \critic where each criterion $c_i$ is accompanied with a set of accepted values $\omega$ as \criterion. For example, The \ca generated the criteria such as Clarity, Efficiency, and more as described in the \tab~\ref{tab:criteria-math} for solving math problems. Each of the Criterion  would be accompanied with a set of accepted values as shown in this Table as an example.

\item The goal of \qa is to quantify each of the suggested criteria to access the task utility $U_t$ for the end user in the following form: \utility, providing us with an idea of the utility of this system for the given task for the end user. In which $s$ represent the task sample and $Q(s|c_i.)$ is the quantifier output for sample $s$ based on the criterion $c_i$.  For example, for one sample of math problem solving, and given the generated criteria shown in \tab~\ref{tab:criteria-math} the solution's Accuracy could be quantified as ``Incorrect'', ``partially correct'' or ``correct''. Eligible quantified values for quantification process are shown in ``Accepted values'' column in \tab~\ref{tab:criteria-math}
\end{itemize}

\medskip
Next we will discuss the datasets and baselines we use to test the work of AgentEval.

\section{Datasets and Solutions}
\label{sec:datasets}

\begin{table}[t]
    \centering
        \caption{Verification Criteria for MathProblems}

    \scalebox{0.56}{
    \begin{tabular}{>{\raggedright\arraybackslash}p{1.9cm}>{\raggedright\arraybackslash}p{6.3cm}>{\raggedright\arraybackslash}p{4.5cm}}
        \hline \hline
        \textbf{Criteria} & \textbf{Description} & \begin{center}
        \vspace{-1em}
            \textbf{Accepted Values}
                    \vspace{-1em}
        \end{center}  \\

        \hline
       Clarity & The ease of understanding the steps, explanations, and language used in the solution. & 
                                       \vspace{-0.5em}
 \begin{enumerate}[label=--,itemsep=0em,parsep=-0.2em,topsep=-4em]
            \item Not Clear   (0)
            \item Moderately Clear (1)
            \item Very Clear (2)
        \end{enumerate} 
       \\[-1em]
        \hline
        Efficiency & The use of optimal methods or approaches to solve the math problem. & 
                                       \vspace{-0.5em}
 \begin{enumerate}[label=--,itemsep=0em,parsep=-0.2em,topsep=-4em]
            \item  Inefficient  (0)
            \item Moderately Efficient (1)
            \item Efficient (2)
 
        \end{enumerate}
    \\[-1em]
        \hline
        Error Analysis & The identification and description of possible errors or misconceptions in the math problem-solving process. &              
                                       \vspace{-0.5em}
 \begin{enumerate}[label=--,itemsep=0em,parsep=-0.2em,topsep=-4em]
            \item  Not Addressed  (0)
            \item  Partially Addressed (1)
            \item Well Addressed (2)
 
        \end{enumerate}
    \\[-1em]
        \hline
        Completeness & Quality of code in terms of efficiency and elegance & 
                                       \vspace{-0.5em}
 \begin{enumerate}[label=--,itemsep=0em,parsep=-0.1em,topsep=-4em]
            \item Incomplete  (0)
            \item Mostly Complete (1)
            \item Complete (2)
        \end{enumerate}
        \\[-1em]
        \hline \hline

    \end{tabular}
    }
    \label{tab:criteria-math}
\end{table}

This section provides an overview of the datasets utilized in our study. Our selection encompasses a variety of datasets, from those based on real-world problems to their simulations and beyond.
The Mathematics Dataset (Section~\ref{sec:math-problems}) is chosen for its widespread usage and comprehensive understanding in the field. It represents complex problem-solving scenarios that are fundamental in evaluating the effectiveness of multi-agent systems.
AlfWorld (Section~\ref{sec:alfworld}) offers a scenario involving multi-turn interactions within a moderately approximated multi-modal environment. This dataset is instrumental in assessing agents' performance in interactive and dynamic settings.

Each dataset plays a critical role in evaluating different aspects of AgentEval's capabilities, from handling complex theoretical problems to navigating real-world scenarios. 
In both tasks, although success is clearly defined, multiple solutions exist for accomplishing the objectives. For example, when solving a math problem, there are various approaches one can take. Similarly, in the Alfworld dataset, which involves household tasks, there are multiple ways to complete them based on how you search for objects and the thinking strategies you employ, among other factors.
An example of Math problem solving and AlfWorld task are shown in Appendix~\ref{sec:task-example}.

\begin{figure*}[ht]
\centerline{%
  \includegraphics[clip, trim=0cm 0cm 0cm 0cm,scale=0.65]{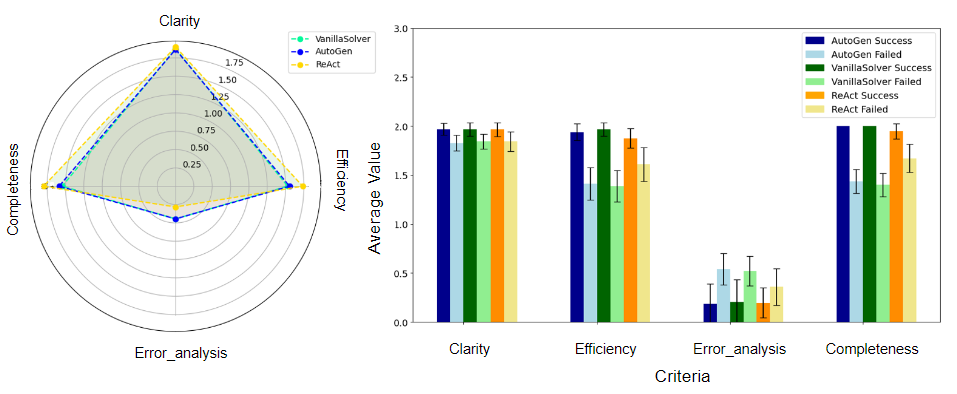}}
\caption{ (a) AgentEval assessment of three different solutions on math problem solving task categorized (b) Same assessment categorized by success and failed cases}
\label{img:math-exp1}
\end{figure*}

\subsection{MATH Problem Solving}
\label{sec:math-problems}
The MATH dataset, originally is a substantial collection of 12,500 challenging mathematics problems from high school competitions~\cite{hendrycks2021measuring}. Each problem comes with a step-by-step solution, enabling models to learn how to generate both derivations and explanations. The dataset covers a wide range of mathematical subjects and is tagged by difficulty levels, offering a nuanced measure of model performance across various aspects of mathematical problem-solving.

This dataset is particularly suitable for testing multi-agent systems for several reason including: 
\begin{enumerate*}[label=(\roman*)]
\item The problems in the MATH dataset are not simple computations but require a deep understanding of mathematical concepts, heuristics, and problem-solving strategies.  
\item Since the dataset includes step-by-step solutions, it allows for the assessment of an agent's ability to learn and reason through a problem, not just its ability to arrive at the correct answer. 
\item The variety of subjects and difficulty levels in the MATH dataset enables a comprehensive evaluation of a system’s versatility and adaptability in different mathematical domains which  is crucial for multi-agent systems that are expected to operate across a range of scenarios.
\end{enumerate*}

Similar to the math problem experimental setup as in \citet{wu2023autogen}, we carry out two experimental evaluations which involves 120 problems from level-5, the most challenging category, and includes 20 problems each from six different categories, of number theory, counting and probability, prealgebra, algebra, intermediate algebra, and precalculus.

%\ahmed{is baselines the best word here. these are different agentic solutions that you are evaluating with AgentEval, not baselines for AgentEval}
\textbf{Solutions:} In establishing a solution for this task, we draw inspiration from the experiments showcased in \cite{wu2023autogen}. We evaluate the proposed methodology by AutoGen~\cite{wu2023autogen}, as well as Langchain ReAct \footnote{\url{https://python.langchain.com/en/latest/index.html}} and a Vanilla solver that employs gpt-4 to tackle the task. These solutions approaches have previously demonstrated promising performance in solving mathematical problems, particularly on the dataset at hand~\cite{wu2023autogen}.
We assess and compare the performance of these three solutions using AgentEval. Fig. \ref{fig:math-example} displays an example of one math problem from prealgebra category as well as the solution created by AutoGen. 
In Section \ref{sec:agenteval-math}, we delve into how AgentEval would perform on math problem solving tasks and how the measured performance with AgentEval correlates with the ground truths.
%\ahmed{what is the takeaway message here. You are basically proposed a new way to measure? does it correlate with ground truth but it is more comprehensive? how do we know that these make sense? If I am building a new Math agent, why should is use AgentEval?}
\subsection{ALFWorld Household Task}
\label{sec:alfworld}

ALFWorld,  presents a set of language-based interactive decision-making tasks within simulated household environments \cite{shridhar2020alfworld}. This benchmark is distinguished by its diversity of tasks, offering a comprehensive platform for testing AI and multi-agent systems. This benchmark is particularly suited for such evaluations because first, ALFWorld is the first interactive parallel environment that aligns text descriptions and commands with physically embodied robotic simulation. It extends two prior works: TextWorld, an engine for interactive text-based games, and ALFRED, a large-scale dataset for vision-language instruction following in embodied environments~\cite{shridhar2020alfred,cote2019textworld}.
The cross-modality framework of this benchmark allows for a variety of embodied tasks with corresponding text-based counterparts, allowing agents to be trained and evaluated in both the language and embodied world. In addition, ALFWorld supports the development of agents that can reason both abstractly and execute actions concretely, mimicking human-like decision-making processes in varying contexts.
Finally, the dataset's inclusion of a wide range of tasks, from household chores to more intricate problem-solving scenarios, provides a comprehensive testbed for evaluating the adaptability and problem-solving capabilities of AI and multi-agent systems. In general, the dataset allows agents to explore, interact, and learn in an abstract language environment before dealing with the complexities of an embodied environment.

\textbf{Solutions:}
As for the solutions to solve ALFWorld Household tasks, similar to \cite{wu2023autogen}, we consider ReAct~\cite{yao2022react} as well as AutoGen with two agents and AutoGen with three agents \cite{wu2023autogen}. 
ReAct is an agent that operates within the ALFWorld environments and is responsible for suggesting plans and executing actions. 
On the other hand, AutoGen Two-Agent System consists of an LLM-backed assistant agent responsible for suggesting plans, and an executor agent responsible for executing actions in the ALFWorld environments. Both ReAct and this solution occasionally struggles with leveraging basic commonsense knowledge about the physical world, which can lead to repetitive errors and getting stuck in loops.
In AutoGen with three agents, a grounding agent is provided just for the sake of critical common sense knowledge whenever the system exhibits early signs of recurring errors.
We assess and compare the performance of these three solutions using AgentEval.
\fig~\ref{alf-example} displays a portion of an example of AlfWorld Householding task solved by AutoGen.

\section{AgentEval Workflow}
\label{sec:workflow}
This Section outlines the workflow of the AgentEval illustrated in Fig.~\ref{fig:agent-eval}.
Next, we will demonstrate how AgentEval works based on 3 different datasets: Math Problems (Sec.~\ref{sec:math-problems}) and AlfWorld (Sec.~\ref{sec:alfworld}).

\subsection{AgentEval for Math Problems}
\label{sec:agenteval-math}

\paragraph{Critic and Quantifier Findings} 
After executing the CriticAgent, we have obtained a set of criteria for validating the results of the mathematical problem presented in Table \ref{tab:criteria-math}. Subsequently, the \qa is tasked with quantifying each criterion based on accepted values. In Figure \ref{img:math-exp1} (a), we present the outcome of \qa i.e., the measured performance of three solutions on this task. This visual representation of the AgentEval output reveals some intriguing insights. Notably, it is evident that Agenteval do not quantify  the three solutions as if they are equally performing well across different criteria. For instance, while all three solutions leverage GPT-4 as the underlying language model, Autogen outperforms ReAct and Vanilla GPT-4 in terms of accuracy. This observation extends to solution completeness and efficiency as well. Conversely, when considering the criterion of Clarity, all three approaches exhibit more competitive performance.

As depicted in this figure, the error analysis range of quantified values differs from other metrics. To gain a better understanding of this criterion, we further scrutinize the results by categorizing them into successful and failed cases, as illustrated in Fig.~\ref{img:math-exp1}~(b).
While AutoGen, Vanilla Solver and ReAct solution are each presented in orange, blue and green respectively, the darker bars represent the performance  on successful cases and lighter bars represent the failed cases performance. The difference between the dark and light bar of each color, verify the AgentEval performance as we expect that each positive criteria should be quantifier higher for successful cases compared to their failed cases. We observe that in most cases, the successful and failed cases are distinguished even with 95\% interval confidence on all the success and failed cases. 

We delve further into the differences between successful cases among the three solutions and failed cases among the solutions. One interesting observation from this Figure is that not all successful cases are identical, and similarly, not all failed cases are the same. The difference between successful cases among the three solutions is smaller than the differences between their failed cases. For instance, Autogen's failed cases exhibit higher efficiency and completeness compared to the Vanilla gpt-4 solver. This observation provides us with valuable additional insights.

\begin{table}[t]
    \centering
        \caption{Verification Criteria for AlfWorld Housholding Tasks. }

    \scalebox{0.56}{
    \begin{tabular}{>{\raggedright\arraybackslash}p{1.9cm}>{\raggedright\arraybackslash}p{6.3cm}>{\raggedright\arraybackslash}p{4.5cm}}
        \hline \hline
        \textbf{Criteria} & \textbf{Description} & \begin{center}
        \vspace{-1em}
            \textbf{Accepted Values}
                    \vspace{-1em}
        \end{center}  \\
        \hline
        Task Understanding & How well the participant was able to comprehend the problem set and follow the task instructions  & 
                              \vspace{-0.5em}
\begin{enumerate}[label=--,itemsep=0em,parsep=-0.2em,topsep=-4em]
            \item Excellent (4)
            \item  Good (3)
            \item Average (2)
            \item Poor (1)
            \item Terrible (0)
          \end{enumerate} 
        \\[-1em]
        \hline
       Plan Making & The ability of the participant to strategize and make a plan for tackling the task. & 
                                       \vspace{-0.5em}
 \begin{enumerate}[label=--,itemsep=0em,parsep=-0.2em,topsep=-4em]
            \item Excellent (4)
            \item  Good (3)
            \item Average (2)
            \item Poor (1)
            \item Terrible (0)
        \end{enumerate} 
       \\[-1em]
        \hline
        Action Decision & The participant's decision-making skills in choosing the right action to perform. & 
                                       \vspace{-0.5em}
 \begin{enumerate}[label=--,itemsep=0em,parsep=-0.2em,topsep=-4em]
            \item Excellent (4)
            \item  Good (3)
            \item Average (2)
            \item Poor (1)
            \item Terrible (0)
 
        \end{enumerate}
    \\[-1em]
        \hline
        Action Execution & How effectively the participant is able to execute the chosen action. &              
                                       \vspace{-0.5em}
 \begin{enumerate}[label=--,itemsep=0em,parsep=-0.2em,topsep=-4em]
            \item Excellent (4)
            \item  Good (3)
            \item Average (2)
            \item Poor (1)
            \item Terrible (0)
 
        \end{enumerate}
    \\[-1em]
        \hline
        Response to Feedback & How well the participant adapts his/her next steps based on the feedback from the environment & 
                                       \vspace{-0.5em}
 \begin{enumerate}[label=--,itemsep=0em,parsep=-0.1em,topsep=-4em]
            \item Excellent (4)
            \item Good (3)
            \item Average (2)
            \item Poor (1)
            \item Terrible (0)
        \end{enumerate}
        \\[-1em]
        \hline
        Correctness of Action &The correctness of the action performed by the participant with respect to the available actions and the current context & 
        \vspace{-0.5em}
  \begin{enumerate}[label=--,itemsep=0em,parsep=-0.1em,topsep=-4em]
            \item Correct  (1)
            \item Incorrect (0)
        \end{enumerate}
        \\
        \hline
                Use of Terminate & Whether the participant uses the 'TERMINATE' command appropriately & 
                \vspace{-0.5em}
  \begin{enumerate}[label=--,itemsep=0em,parsep=-0.1em,topsep=-4em]
            \item Appropriate  (1)
            \item Inappropriate (0)
        \end{enumerate}
        \\[-1em]
         \hline \hline
    \end{tabular}
    }
    \label{tab:criteria-alf}
\end{table}

\subsection{AgentEval for AlfWorld}
\paragraph{Critic and Quantifier Finding} 
In this section, we provide an example of AgentEval applied to the AlfWorld Householding task, as mentioned in Sec.~\ref{sec:agenteval-math} in which real-world household environments is emulated through textual interfaces \cite{shridhar2020alfworld}. When running the \ca on this task, it identified specific criteria such as ``Task understanding'', ``Plan making'' and ``Response to Feedback'' as outlined in Tab.~\ref{tab:criteria-alf}. We consulted researchers deeply involved with these tasks, and their expertise confirmed that these criteria are critically relevant and significant similar to~\cite{li2023collaborative}.
For example, given that these tasks are language-based and require interactive decision-making, an agent in ALFWorld is tasked with high-level objectives, such as placing a hot apple in the fridge, and must navigate and interact with a simulated household environment to achieve these objectives. Therefore, criteria displayed in Tab.~\ref{tab:criteria-alf} satisfy the assessment of this task. While the criteria are pretty self-descriptive, about the criterion ``Use of TERMINATE'' We note that the agent is prompted to use the term ``TERMINATE'' upon task completion, which is closely correlated with task success.

\begin{figure*}[ht]
\centerline{%
  \includegraphics[clip, trim=0cm 0cm 0cm 0cm,scale=0.6]{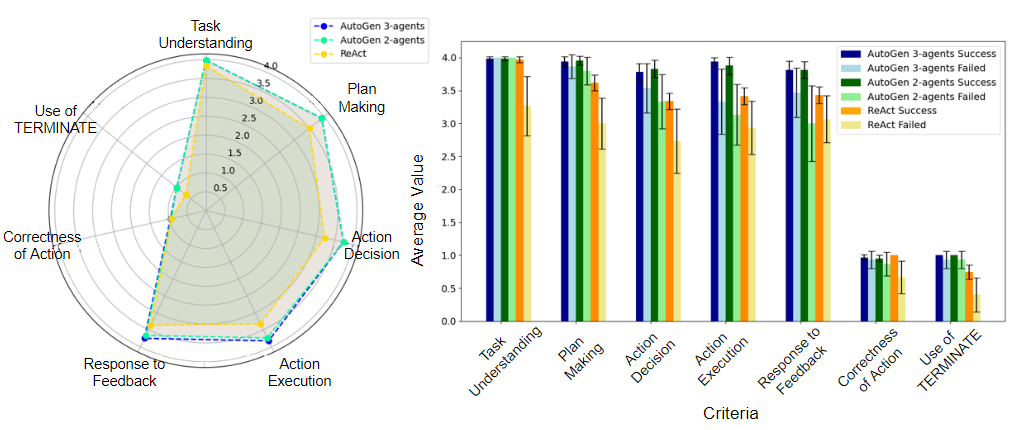}}
  \caption{(a) AgentEval assessment of three different solutions on AlfWorld Householding Task (b) Same assessment categorized by success and failed cases.}
\label{img:alf-exp1}
\end{figure*}

Following the extraction of a set of criteria as detailed in Tab \ref{tab:criteria-alf}, these criteria are passed to the QuantifierAgent for quantification on each sample. Figure \ref{img:alf-exp1} presents the results for three introduced solutions: AutoGen with 2 agents, AutoGen with 3 agents, and ReAct, on the 134-test set from \cite{wu2023autogen}. On the left side of Fig.~\ref{img:alf-exp1}, a Spider Figure illustrates the performance of these three solutions across all criteria. It is important to note that all criteria, except ``Use of TERMINATE'' and ``Correctness of Action'' employ a five-level grading system, while these two criteria are binary. From this figure, it is evident that ReACT performs notably worse across all criteria, while AutoGen with 2 agents and 3 agents demonstrate competitive performance. Notably, AutoGen with an additional common-sense grounding agent slightly outperforms others, particularly in the areas of Response to Feedback and Action Execution.
Additionally, the barplot on the right side of \fig~\ref{img:alf-exp1} categorizes the 134 games into two groups: failed and successful, displaying the quantifier performance for each subgroup. Similar to \fig~\ref{img:math-exp1}, darker colors represent performance in successful cases for each solution, while lighter colors represent performance in failed cases. AutoGen 3-agent, AutoGen 2-agent, and ReAct are represented by blue, green, and orange, respectively. For most criteria, the distinction between failed and successful cases is clear, even within a 95\% confidence interval. However, for certain criteria, such as ``Task understanding'' all solutions, whether they failed or succeeded, exhibit very similar performance. This could be interpreted as either (1) all solutions have a good understanding of the task, even if they fail to complete it, (2) this criterion may be redundant, as it does not provide additional information among these three solutions or (3) the \qa is unable to score the criterion in a meaningful way. We refrain from concluding which criteria are most suitable for this specific task. Instead, we emphasize the importance of conducting a more in-depth analysis of performance beyond success rates, tailored to one's goals and application requirements.

\section{AgentEval Robustness Analysis and In-depth Discussion}
\label{sec:results}

This section presents the results of the analysis of how robust AgenEval is. First, we inspect if the list of criteria can be solely extracted from the task description (task-based criteria), and how the list of criteria can be changed by adding failed and successful samples from the data. Where we played with varies sample size to check its effect of the final list of criteria (Section~\ref{sec:c-robust}).
Second, we focus on how can we estimate the robustness of the \qa (Section~\ref{sec:q-robust}). We note that all the experiments reported in the paper are conducted with the temperature set at 0.
Next, we will present our analysis using the MATH Problems dataset.

\begin{figure*}[!t]
\centerline{%
\includegraphics[width=1\textwidth]{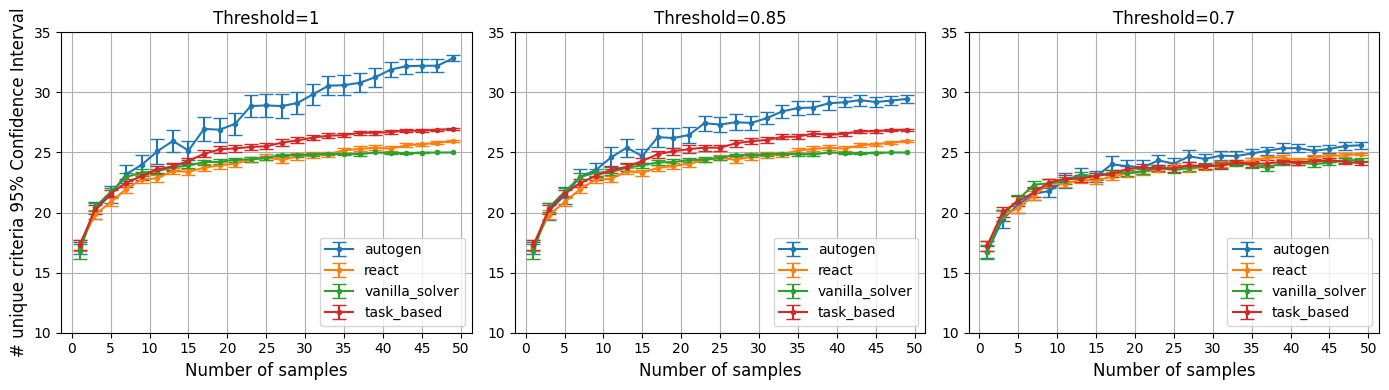}}
\caption{Task based criteria vs solution based criteria for Math problems. show the 95\% interval at each step}
\label{fig:task-vs-solution}
\end{figure*}

\subsection{Task-based vs Solution-based criteria}
\label{sec:c-robust}

\paragraph{General Hypothesis} We execute the CriticAgent using two distinct methods. The first method involves the Agent generating criteria solely based on the provided task description, which we refer to as ``task-based'' criteria. On the other hand, the CriticAgent could potentially derives criteria not only from a task description but also from examples of task solutions so called as ``solution-based'' criteria. In this context, our objective is to examine whether this approach leads to variations in the criteria formulated by agents. We believe this investigation is important to have a more clear vision of what criteria necessitate for having a promising assessment.

A solution to a mathematical problem, might probably satisfy criteria such as accuracy and clarity in any case, independent of what the solution is. However, when additional tools are being utilized to solve the problems, such as coding to solve math problems, additional criteria like `Code Efficiency' may be introduced to the set of criteria. If one never considered solving the problem with a specific solution method like coding, they might not initially include such criterion. In summary, depending on whether the \ca receives only a task description or both a task description and examples of solutions, we classify the criteria as either ``task-based'' or ``solution-based''.
Additionally, it is important to analyze whether the solution-based criteria overlap across different solutions and to what extent different solutions share these criteria.

To compare the differences between task-based and solution-based criteria, Fig.~\ref{fig:task-vs-solution} displays the number of unique criteria extracted for mathematical problem solving in task-based mode and three different solution-based approaches i.e., when the solutions come from AutoGen, ReAct and Vanilla Solver. To keep the balance between computational costs and analyzing the robustness, we conducted 50 runs of the CriticAgent with different seeds. Subsequently, for $N=50$ iterations, we randomly selected $M \in [1,50]$ samples ($M$ is shown on the x-axis of Fig.~\ref{fig:task-vs-solution}) and present the average number of unique extracted criteria along with its 95\% confidence interval after repeating this process 50 times. We note that because we obtained results from the \ca in 50 iterations in total, the confidence intervals become smaller when $M$ get closer to the maximum number of samples i.e., $50$.

When examining the criteria, we have identified instances where certain criteria are quite similar but are expressed differently. These are essentially metrics that convey the same concept but are phrased with slight variations. In Table \ref{tab:sim_criteria}, we provide examples of such similarities along with their descriptions. In order to gain a deeper insight into the results presented in Figure \ref{fig:task-vs-solution}, we suggest consolidating these closely related criteria to determine the total number of unique criteria once again. This approach serves two purposes:
1. It enhances our understanding of the actual number of unique criteria that have been extracted.
2. It allows us to assess whether the repetitiveness and redundancy of criteria differ between solution-based and task-based criteria.
By doing so, we can gain a better grasp of the data and draw more meaningful conclusions from our analysis.

\begin{table}[t]
    \centering
    \caption{Pairs of similar criteria extracted for Math problem solving task.}
    \scalebox{0.72}{
    \begin{tabular}{p{10cm}}
    \hline\hline
      - Problem Difficulty: The complexity of the math problem that has been solved. 
 \\   - Problem Complexity: The level of difficulty of the problem.\\ \hline
      - Innovativeness: The novelty and creativity in the approach to solve the problem 	 
   \\ - Innovation: The ability to solve a problem using a unique or creative method not commonly known.\\ \hline
      - Time Taken: The time taken to solve the problem. 	 
      \\ - Time to Completion: The amount of time taken to solve the problem completely \\\hline
       - Understandability: The clarity and ease of comprehension of the solution provided. 	
    \\ - Readability: How easy it is to comprehend the provided solution. \\\hline\hline
    \end{tabular}}
    \label{tab:sim_criteria}
\end{table}

\begin{figure*}[t]
\centerline{%
\includegraphics[width=1.02\textwidth]{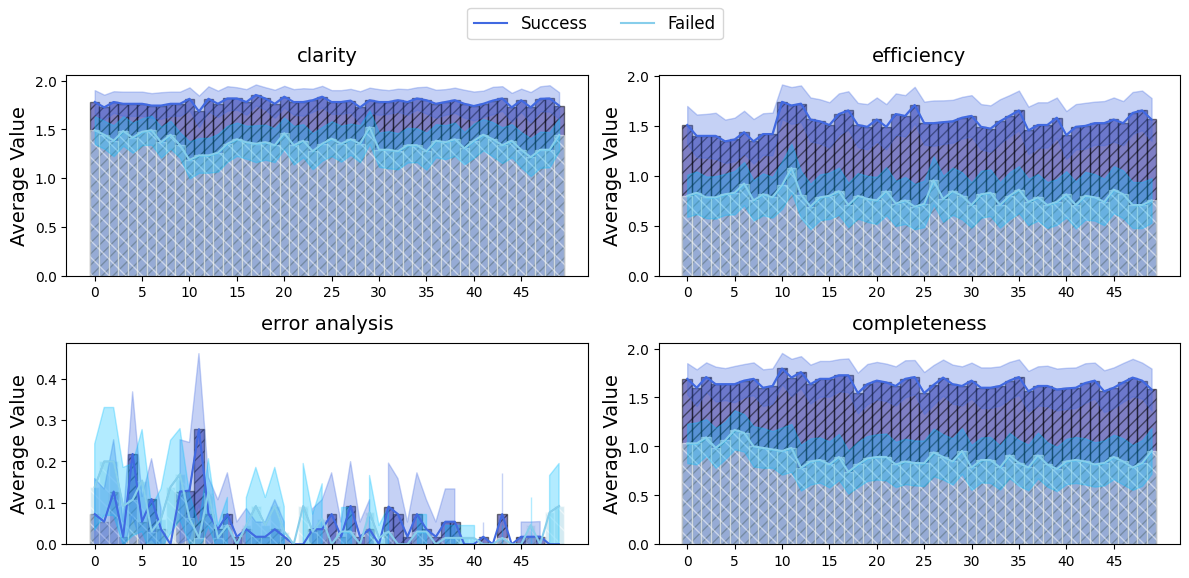}}
\caption{Quantifier Robustness on criteria of Math Problem Solving problem. Each bar represent the average performance of success (dark blue "//") and failed (light blue ``\textbackslash\textbackslash'') cases and 95\% interval on each set is  shaded across the average point.  The two plots are overlaid.}
\label{fig:q-robust-first}
\end{figure*}

In order to consolidate similar criteria, we draw inspiration from previous work \cite{liu2022semantic,vahtola2022not,reimers2019sentence} which demonstrated that utilizing pre-trained language models fine-tuned for paraphrasing and semantic similarity can yield high performance in numerous downstream NLP tasks. Additionally, we employ a fine-tuned pre-trained language model specifically designed for paraphrasing, known as the Hugging Face Paraphrase MiniLM \footnote{\url{https://huggingface.co/sentence-transformers/paraphrase-MiniLM-L6-v2}}.

Our approach begins by encoding each criterion's title and its description, followed by measuring pairwise similarity between all available criteria within our experiments. Subsequently, by employing a specified threshold value denoted as $\tau$, we classify pairs with higher cosine similarity between the embedded representations of each criterion pair as one and select one of them as the representative for that pair. This strategy is commonly employed in various NLP downstream tasks.

In \fig~\ref{fig:task-vs-solution}, we illustrate the outcomes of the number of unique extracted criteria using different threshold values, namely  0.7, 0.85, and 1. A threshold of 1 implies that no criteria are filtered out.

\paragraph{Summary} 
In this section, we delved into various inputs and methods for extracting criteria. Our exploration compared the outcomes of task-based criteria, derived solely from task descriptions, with those of solution-based criteria, where the \ca is exposed to both examples of solutions and the task description. We observed that solution-based methods produce a greater diversity of criteria compared to task-based methods. Furthermore, the diversity in the unique number of criteria varied even within solution-based methods, influenced by the model's level of creativity. Additionally, we noticed a tendency for certain criteria to recur when running the \ca multiple times. To address this, we suggest implementing consolidation techniques, such as merging synonymous terms, to eliminate redundant criteria."

\subsection{Quantifier Agent Robustness}
\label{sec:q-robust}

\paragraph{General Hypothesis}
Here, we aim to investigate the robustness of the \qa when applied repeatedly to the same set of criteria. Our goal is to assess the consistency of the results when quantifying the same set of criteria multiple times. This is of utmost importance as we expect the behavior of the quantifier to be stable and relatively free from noise when provided with a single sample and a fixed set of criteria. This stability is crucial for us to have confidence in the results. Additionally, this analysis can help us identify and filter out criteria that may not be sufficiently stable for reliable use.

To achieve this, we selected a specific subset of criteria related to mathematical problems, as detailed in Table \ref{tab:criteria-math}, and conducted 50 runs of the quantifier agent on the 120 problems described in Section \ref{sec:math-problems}. Our expectation is to observe consistent quantified performance for each of the criteria. In \fig~ \ref{fig:q-robust-first}, we present the distribution of quantified performance across 50 runs for both successful and failed cases, focusing on the five selected criteria. A consistently horizontal performance trend indicates greater robustness in the quantifier, whereas more fluctuations in the figure suggest less robustness and a noisier performance of the agent.

As shown in the results, for four out of the five generated criteria, we consistently observe steady performance. Not only do the success cases consistently outperform the failed cases, but their performance also falls within a similar range across runs. However, when it comes to the ``error analysis'' criterion, we observe a more variable performance of the quantifier. It does not consistently predict one group (success or failed) to perform better than the other, and the quantifier's performance varies across different runs. This suggests that the AgentEval tool may not exhibit promising robustness for this particular criterion. The underlying issues could be either the criterion itself lacks clarity and appropriateness for the task, or the \qa struggles to quantify this criterion effectively. In either case, it is advisable to either modify or eliminate this criterion to enhance trustworthiness and reliability.

\begin{figure}[!tb]
\centerline{%
\includegraphics[width=0.47\textwidth]{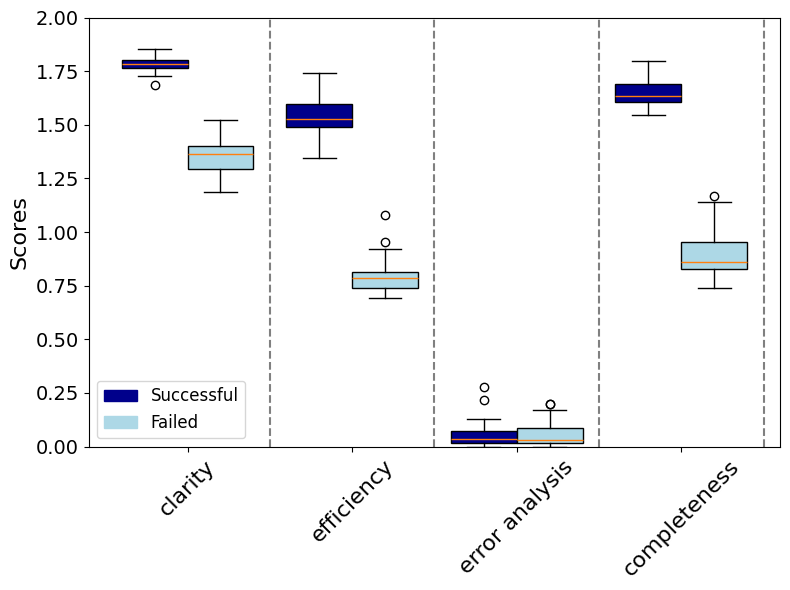}}
\caption{Quantifier Robustness - Distribution of QuantifierAgent output on AutoGen results on 120 Math problems on Success (dark blue) and Failed (light blue) cases on terms of different criteria. The distributions demonstrates the same results as in Fig \ref{fig:q-robust-first}.}
\label{fig:q-box-plot}
\end{figure}

Furthermore, we present the distribution of quantified values in \fig~\ref{fig:q-box-plot} using box plots, illustrating the distribution of quantifier values for both failed (dark blue) and successful cases (light blue) across all criteria. The box plots display the first and third quartiles of the distribution as well as the median. In this figure, robust criteria should exhibit a narrower range of quantifier performance (narrower box plots), and it should be easy to distinguish between the dark and light box plots for each criterion.

Consistently with our previous observations, all four criteria, except ``error analysis'' allow for easy differentiation between successful and failed cases. Additionally, some criteria prove to be more robust compared to others. For example, accuracy displays a narrower range of distribution, while clarity in failed cases covers a wider range. We believe that such an analysis of the quantifier agent's performance will yield valuable insights for enhancing reliability, trustworthiness, and explainability in performance evaluation.

\paragraph{Summary}
We recognize the importance of thoroughly investigating the robustness of each criterion in quantification studies. This analysis is crucial as it sheds light on the stability of each criterion. Moreover, when ground truths are available, such as in cases of success versus failure, they provide a benchmark to validate our assessments. Additionally, it's important to acknowledge that not all criteria exhibit the same level of robustness. This variability demands careful consideration during evaluations, especially given the non-deterministic nature of LLMs. Such awareness is essential to ensure the reliability and accuracy of our assessments in the dynamic field of LLMs.

\subsection{\qa Verification}
\label{sec:q-verify}

To assess the accuracy of quantifying each criterion, it is essential to verify the quantification process. Ideally, we would like to validate this process by comparing it with known pairwise samples, where we have definitive knowledge that for a given criterion $C$,  sample $A$ is superior compared to sample $B$. The correct quantification should align with this knowledge. However, as the use of LLM-powered applications continues to expand daily, obtaining annotated data for many tasks is often impractical, if not impossible. Therefore, we propose employing synthetically altered versions of the samples to obtain the knowledge required for this verification.

Let us assume that we have an alternative disturbed version of sample $A$, which is called $A'$. Assuming that the original sample $A$ outperforms the one with injected noise $A'$, we anticipate that the criteria that assess sample quality will assign higher values to the original sample compared to the noisier variant in the same case. To carry out this validation, we conducted experiments involving mathematical problems. We introduce random noise into the solutions by removing a certain percentage of the solution sentences from Autogen's results for the math problem solving dataset. For criteria such as ``completeness'' or ``clarity'', we expect to observe greater completeness or clarity in the original solution as opposed to the one missing a portion of the solution.

\begin{figure}[t]
\centerline{%
\includegraphics[width=0.5\textwidth]{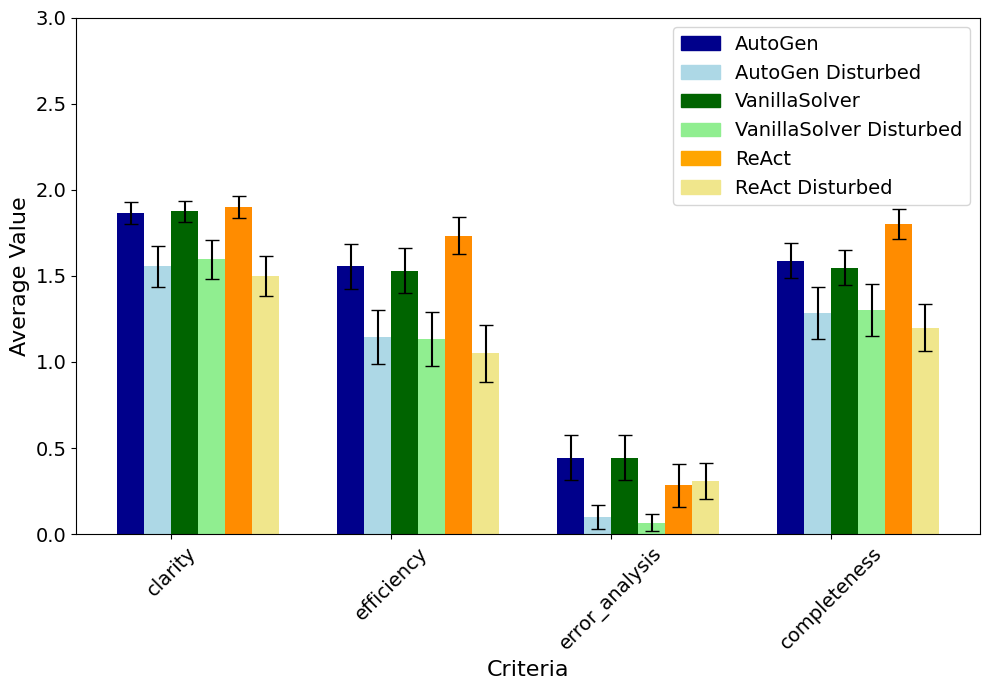}}
\caption{Quantifier Verification on original set of solutions as well as the disturbed solutions on Math Problem Solving dataset.}
\label{fig:q-verify-disturb}
\end{figure}

In our study, our goal is to assess the \emph{QuantifierAgent's} ability to capture these distinctions between a known better solution and a worse one. We generated disturbed versions of solutions by randomly removing 25\% of the sentences and running the quantifier over the noisy solutions. The results of these experiments are presented in \fig~\ref{fig:q-verify-disturb}. As depicted in this figure, the  criteria that captures the quality of the solutions such as ``clarity'' and ``completeness'' of the disturbed solutions decreased compared to the original ones. This observation helps establish confidence in the performance of \qa.

\section{Conclusions and Future Work}
\label{sec:conclusions}

The rapid development of open-source libraries aiming to simplify the creation of Language Model Models (LLM)-powered agentic solutions for various user-centric tasks has facilitated the rapid growth of such applications. However, meeting end-users' expectations and requirements for these applications is paramount, underscoring the importance of assessing the utility they provide.
Directly evaluating agentic systems presents challenges, as current approaches often rely on end-to-end success metrics alone. However, understanding user interaction with an application entails more than just task success. Given the diverse range of tasks requiring automation, a scalable and flexible methodology is essential for evaluating these applications effectively.

In this work, we introduce the AgentEval framework, designed to swiftly gauge the utility of LLM-powered agentic applications for end-users. AgentEval aims to assess the alignment between application behavior and user goals, providing developers with insights into areas for improvement. The framework leverages recent findings suggesting LLMs as a scalable and cost-effective alternative to human evaluations for open-ended tasks. AgentEval consists of two agents: \ca suggests criteria based on task descriptions and suggested solutions, while \qa verifies how well the solutions align with these criteria. This framework is customizable, adaptable, and can operate in various modes, employing combinations of LLMs, human inputs, and tools.
We believe that AgentEval's utility extends beyond immediate performance verification. It can uncover new system capabilities over time and adapt to changes in user needs or developer requirements. 

In summary, our contributions include defining task utility, introducing the AgentEval framework, and conducting a robust analysis of its performance across various datasets and solutions. AgentEval represents a significant step towards evaluating and optimizing LLM-powered applications to better serve end-users.

\newglossaryentry{System}{name={System}, description={is the multi-agent system that is developed using AutoGen (similar type of libraries}}

\newglossaryentry{example}{name={example}, description={A sample term}}

\printglossary

\bibliography{acl}
\bibliographystyle{acl_natbib}

\appendix
\newpage
\section{Appendix}
\subsection{Task Examples}
\label{sec:task-example}
In Figure \ref{fig:math-example} and \ref{alf-example}, we display an example of Math problems and AlfWorld Householding tasks with AutoGen's solution for these problems.

\begin{figure*}[!h]
\begin{tcolorbox}[colback=gray!5!white,colframe=gray!75!black]
{
  "problem": "How many positive and negative integers is $12$ a multiple of?"
  
  "level": "Level 5"
  
  "type": "Prealgebra"

    "response\_with\_ans": 
    
  "messages": [
  
    \{     
    
    "content": "How many positive and negative integers is $12$ a multiple of?",
      
      "role": "user"
    
    \},
    
    \{
    
    "content": "To determine how many positive and negative integers 12 is a multiple of, we need to find the divisors of 12. Then we will double the number of divisors, as each positive divisor has a corresponding negative divisor.
      Let's use a Python code snippet to find all the divisors of 12 and count them.",
      \begin{lstlisting}[language=Python]
num = 12
divisors = [i for i in range(1, num + 1) if num \% i == 0]
positive_divisors_count = len(divisors)
negative_divisors_count = positive_divisors_count
total_divisors_count = positive_divisors_count + negative_divisors_count
print(f"Positive divisors of 12: {divisors}")
divisors_negative = [-x for x in divisors]
print(f"Negative divisors of 12: {divisors_negative}")
print(f"Total divisors count: {total_divisors_count}")
\end{lstlisting}

    Execute this code snippet to find the divisors of 12 and count how many positive and negative integers 12 is a multiple of.
    
      "role": "assistant"
  
    \}
 
  ],
  
  "time": 17.360238790512085

}
 
\end{tcolorbox}
\caption{An example of Math Problem solving sample with solution generated by AutoGen}
\label{fig:math-example}
\end{figure*}

\begin{figure*}[t]
\begin{Verbatim}[fontsize=\footnotesize, breaklines]
{
    { "content": "Perform actions and interact with a household to solve a task. At the beginning of your interactions, you will be given the detailed description of the current environment and your goal to accomplish. For each of your turn, you should choose from two actions: "think" or     "Action". If you choose "think", you should first think about the current condition and plan for your future actions, and then output your action in this turn. Your output must strictly follow this format:"think: your thoughts."
    "Action: your next action\\n"; If you choose "ACTION", you should directly output the action in this turn. Your output must strictly follow this format:"ACTION: your next action".
    After each of your turn, the environment will give you immediate feedback based on which you should plan your next few steps. if the envrionment output "Nothing happened", that means the previous action is invalid and you should try more options.
    Reminder: 
    1. the action must be chosen from the given available actions. Any actions except provided available actions will be regarded as illegal.
    2. Take the target household immediatly after you find it.
    3. Reply 'TERMINATE' only under two circumstances: a). The task has given you clear instructions to return 'TERMINATE' b). The task seems insolvable.\\n Here are two examples.\nYou are in the middle of a room. Looking quickly around you, you see a cabinet 13, a cabinet 12, a cabinet 11, a cabinet 10, a cabinet 9, a cabinet 8, a cabinet 7, a cabinet 6, a cabinet 5, a cabinet 4, a cabinet 3, a cabinet 2, a cabinet 1, a coffeemachine 1, a countertop 1, a diningtable 1, a drawer 1, a fridge 1, a garbagecan 1, a microwave 1, a shelf 3, a shelf 2, a shelf 1, a sinkbasin 1, a stoveburner 4, a stoveburner 3, a stoveburner 2, a stoveburner 1, and a toaster 1.\nYour task is to: put a hot apple in fridge.\n",
        "role": "user"},
    {   "content": "THOUGHT: The task is to put a hot apple in fridge. So i need to first find an apple and heat it. First I need to find an apple. An apple is more likely to appear in diningtable (1), coffeetable (1), fridge (1), drawer (1), cabinet (1-13), garbagecan (1). I can check one by one, starting with fridge 1.\n ACTION: go to diningtable 1\n",
        "role": "assistant"},
    {   "content": "On the diningtable 1, you see a apple 1, a bread 1, a fork 3, a fork 2, a fork 1, a mug 2, a peppershaker 3, a plate 2, a pot 1, a soapbottle 3, a spatula 1, a spoon 3, a spoon 2, a spoon 1, a winebottle 3, and a winebottle 2.\n",
        "role": "user"},
    {   "content": "ACTION: take apple 1 from diningtable 1\n",
        "role": "assistant"},
    {   "content": "You pick up the apple 1 from the diningtable 1.\n",
        "role": "user"},
    {   "content": "THOUGHT: I noticed that there is a microwave 1 in the room, i will use it to heat the apple.\n ACTION: go to microwave 1\n",
        "role": "assistant"},
    ...
    ...
 
    {   "content": "You open the microwave 1. The microwave 1 is open. In it, you see a egg 1.",
        "role": "user"
    },
    {   "content": "ACTION: heat mug 2 with microwave 1",
        "role": "assistant"},
    {    "content": "You heat the mug 2 using the microwave 1.Task success, now reply TERMINATE\n",
        "role": "user"},
    {   "content": "TERMINATE",
        "role": "assistant"},
    {   "content": "TERMINATE",
        "role": "user"}
}
\end{Verbatim}
\caption{An example of AlfWorld Householding task from ``Pick, Heat, then Place'' category.}

\label{alf-example}

\end{figure*}

\label{sec:appendix}

\end{document}